\documentclass{article}

    \PassOptionsToPackage{numbers, compress}{natbib}
\usepackage[sglblindworkshop, final]{ai4nextg_neurips_2025}

\usepackage[utf8]{inputenc} % allow utf-8 input
\usepackage[T1]{fontenc}    % use 8-bit T1 fonts
\usepackage{hyperref}       % hyperlinks
\usepackage{url}            % simple URL typesetting
\usepackage{booktabs}       % professional-quality tables
\usepackage{amsfonts}       % blackboard math symbols
\usepackage{nicefrac}       % compact symbols for 1/2, etc.
\usepackage{microtype}      % microtypography
\usepackage{xcolor}         % colors
\usepackage{amsmath}
\usepackage{graphicx}
\usepackage{algorithm}
\usepackage{algpseudocode} 
\usepackage{subcaption}
\usepackage{multirow}

\title{VLF-MSC: Vision-Language Feature-Based Multimodal Semantic Communication System}

\author{%
  Gwangyeon~Ahn\thanks{These authors contributed equally.}, \quad 
  Jiwan~Seo\footnotemark[1],  \quad 
  Joonhyuk~Kang\thanks{Corresponding author: J. Kang (jkang@kaist.ac.kr).}\\
  Department of Electrical Engineering \\
  Korea Advanced Institute of Science and Technology (KAIST)\\
  Daejeon, 34141, South Korea\\
  \texttt{hihi22@kaist.ac.kr}, \ \ \ \texttt{jeewan0516@kaist.ac.kr}, \ \ \ \texttt{jkang@kaist.ac.kr} \\
}

\begin{document}

\maketitle

\begin{abstract}
We propose Vision-Language Feature-based Multimodal Semantic Communication (VLF-MSC), a unified system that transmits a single compact vision-language representation to support both image and text generation at the receiver. Unlike existing semantic communication techniques that process each modality separately, VLF-MSC employs a pre-trained vision-language model (VLM) to encode the source image into a vision-language semantic feature (VLF), which is transmitted over the wireless channel. At the receiver, a decoder-based language model and a diffusion-based image generator are both conditioned on the VLF to produce a descriptive text and a semantically aligned image. This unified representation eliminates the need for modality-specific streams or retransmissions, improving spectral efficiency and adaptability. By leveraging foundation models, the system achieves robustness to channel noise while preserving semantic fidelity. Experiments demonstrate that VLF-MSC outperforms text-only and image-only baselines, achieving higher semantic accuracy for both modalities under low SNR with significantly reduced bandwidth.
\end{abstract}

\section{Introduction}

Artificial Intelligence (AI) and Machine Learning (ML) are rapidly advancing and increasingly integrated into wireless communication systems to support data-intensive applications such as augmented/virtual reality, autonomous driving, and massive Internet of Things (IoT) networks~\citep{kim2018communication, eldar2022machine}. 
These emerging services generate diverse multimodal data that must be efficiently transmitted over noisy and bandwidth-limited channels.
To meet these demands, a paradigm shift is required from conventional data-centric communication towards semantic communication (SC), which focuses on transmitting meaning rather than exact data representations~\citep{9679803, 9955525, 9955312}.

Recent advances in large-scale language models (LLMs) and generative models have greatly improved multimodal understanding, making them a strong foundation for SC systems~\citep{wang2024large, guo2025large}. However, traditional modality-specific strategies face key limitations. Text-based SC~\citep{DeepSC1} cannot capture fine-grained visual details, while image-based systems demand high bandwidth and may introduce semantic ambiguity. Moreover, when receiver requirements change, separate transmissions or extra mechanisms are needed since each modality is processed independently, incurring additional overhead.

Vision-language models (VLMs) open a promising path to address these challenges and have already been integrated into SC frameworks~\citep{11032085, Liang_2025_CVPR}. A typical VLM encodes images and text into a single fused representation that bridges both domains. Such unified vision-language features capture rich semantic interactions across modalities, improving tasks like captioning, retrieval, and question answering~\citep{clip, dalle, blip2}. Building on these advances, we propose a multimodal semantic communication system that leverages this unified representation for efficient transmission.

We introduce Vision-Language Feature-based Multimodal Semantic Communication (VLF-MSC), a system that transmits a compact Vision-Language Feature (VLF) to support multimodal reconstruction at the receiver. Instead of sending modality-specific encodings (e.g., separate text or image descriptions), the transmitter extracts a modality-agnostic semantic feature using a VLM and sends it over the channel. The receiver then employs two decoders: a text decoder (a decoder-based LLM) to generate captions, and an image decoder (diffusion model) to generate semantically aligned images, both guided solely by the received feature. This design enables a single transmitted representation to support multiple modalities, allowing the receiver to flexibly obtain the desired output without retransmission. By leveraging powerful pre-trained models, VLF-MSC remains robust to channel impairments and preserves semantic fidelity of the outputs. Experiments demonstrate that VLF-MSC achieves superior transmission efficiency and preserves semantic similarity of both text and image generations under noisy channels, outperforming conventional approaches. Our contributions are summarized as follows:
\begin{itemize}
    \item We transmit a single VLF extracted by the VLM, using it as the context for both a decoder LLM and a diffusion image decoder at the receiver. This modality-agnostic design eliminates the need for modality-specific streams and retransmissions.
    \item As input image resolution increases, the transmit budget remains resolution-invariant. The transmitter maps the VLF directly onto the transmitted symbols, and the receiver performs prompt-free generation. This yields a lightweight, deployable SC pipeline.
    \item VLF-MSC achieves higher perceptual fidelity and stronger image-text semantic alignment than baseline models, and preserves textual meaning more faithfully than text-centric baselines at low SNRs.
\end{itemize}

\section{Related Work}
\paragraph{Multimodal Representation Learning via VLMs}
VLMs have achieved breakthroughs in connecting vision and language through large-scale pre-training on image-text data~\citep{11032085}. Models such as CLIP~\citep{clip}, DALL·E~\citep{dalle}, and BLIP-2~\citep{blip2} show that visual and textual signals can be embedded into a shared semantic space. CLIP aligns image and text embeddings via contrastive learning, enabling zero-shot recognition and retrieval, while DALL·E demonstrates generative cross-modal capabilities by producing images from text. BLIP~\citep{blip1} and BLIP-2 bridge frozen vision transformers and language models through a lightweight Q-Former that extracts informative query embeddings. Extensions like BLIP-Diffusion~\citep{blip_diffusion} further couple these embeddings with diffusion models for subject-driven image generation, highlighting the versatility of vision-language features.

\paragraph{Multimodal Semantic Communication}
Research in semantic communication has recently expanded to multimodal settings~\citep{gu2023semantic, 10670195}. Early approaches transmitted intermediate text captions to convey the intent of visual data~\citep{10446638}. \citet{LaMoSC} instead used a large language model at the receiver to interpret visual embeddings directly. More recent works propose sending joint latent representations of image and text with generative decoders for reconstruction~\citep{DiffusionSC2}, while diffusion-based methods improve robustness under noise. Img2Img-SC~\citep{LatentSC} combines BLIP-2 text embeddings with a compact visual latent code, enabling bidirectional decoding. These studies demonstrate that pre-trained VLMs are powerful tools for multimodal semantic communication. Our work advances this line by transmitting a \emph{single} VLF that supports both text and image decoding, simplifying transceiver design and avoiding alignment overhead. Unlike prior multimodal SC approaches that rely on either modality-specific codes or joint embeddings with extra alignment, our framework achieves robust multimodal communication from a unified semantic representation, enhancing both efficiency and scalability across tasks.
\begin{figure}[!t]
\begin{center}
\centerline{\includegraphics[width=\linewidth]{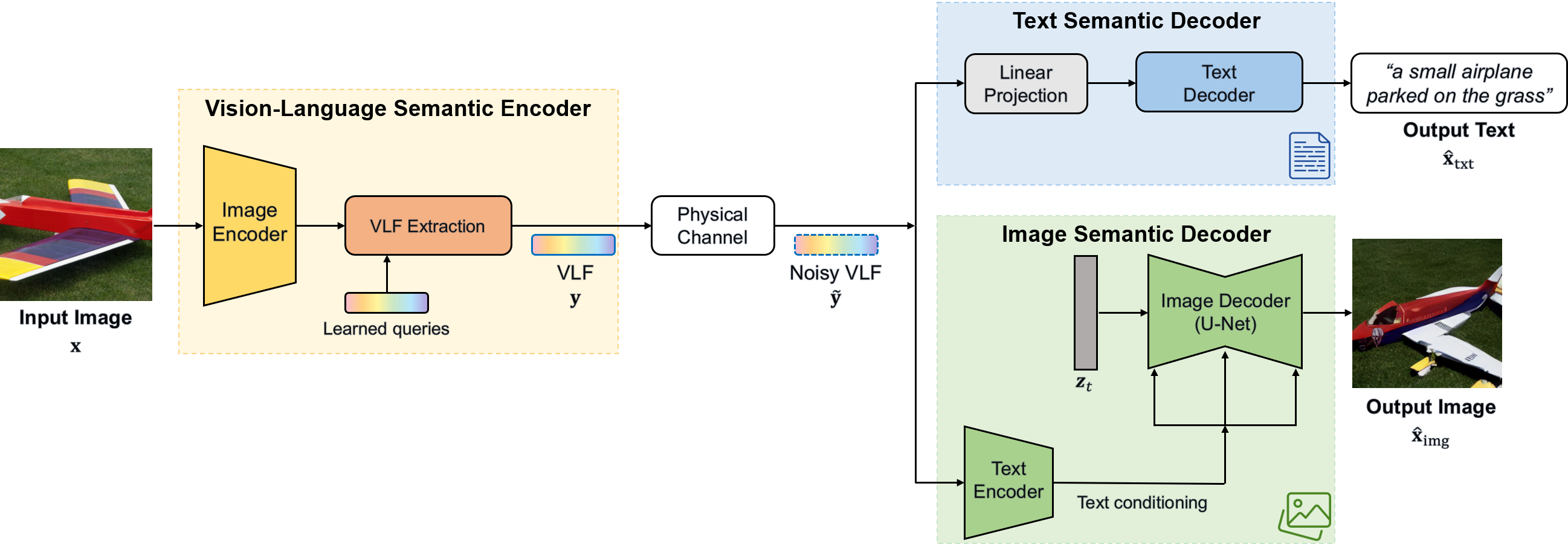}}
\caption{
Overview of the proposed VLF-MSC framework.
The transmitter encodes an input image $\mathbf{x}$ into a compact VLF $\mathbf{y}$ using an image encoder and Q-Former, which is transmitted over a noisy channel.
At the receiver, the received VLF $\tilde{\mathbf{y}}$ jointly conditions the text and image semantic decoders to generate both the textual output $\hat{\mathbf{x}}{\text{text}}$ and the visual output $\hat{\mathbf{x}}{\text{img}}$, enabling multimodal communication without separate transmissions.
}
\label{fig:1}
\end{center}
\vskip -0.2in
\end{figure}

\section{Proposed Method}

Figure~\ref{fig:1} illustrates the proposed VLF-MSC system, which embeds vision-language semantic features within a communication framework for multimodal data transmission. The system consists of three main components: the transmitter, the physical channel, and the receiver. At the transmitter, a vision-language semantic encoder processes the source image and outputs a compact semantic feature. Specifically, a pre-trained image encoder extracts visual features, and a Querying Transformer (Q-Former) produces a set of query embeddings that encode both visual and semantic information. This vision-language feature (VLF) is transmitted over the wireless channel, modeled as a noisy link that perturbs the features. At the receiver, a multimodal semantic decoder reconstructs the target modality from the noisy VLF. Two decoders are implemented: one for text and one for images, both conditioned on the same received VLF. Thus, the transmitted VLF serves as the sole context for both decoders, enabling flexible modality selection while maintaining efficiency. In other words, a single transmitted semantic feature can drive textual description, image reconstruction, or both (without requiring separate streams, highlighting the communication efficiency of our design).

\subsection{Vision-Language Semantic Encoder}
The transmitter’s semantic encoder, based on BLIP-2~\citep{blip2}, converts a high-dimensional image into a compact vision-language feature suitable for transmission. Given an input image $\mathbf{x} \in \mathbb{R}^{3\times H \times W}$, the encoder produces $N$ query vectors of dimension $d$, forming $\mathbf{y} \in \mathbb{R}^{N \times d}$ as the semantic representation.

\paragraph{Image Encoder}
Given an input image $\mathbf{x}$, we employ a frozen Vision Transformer (ViT)~\citep{vit} to extract high-level visual features. The image is partitioned into non-overlapping patches, linearly projected, and processed by transformer blocks to yield patch-level feature vectors. This stage remains frozen during training, simplifying optimization and stabilizing alignment with language features.

\paragraph{VLF Extraction via Q-Former}
To integrate visual features with language semantics, we employ Q-Former from BLIP-2. The Q-Former acts as a multimodal encoder and an information bottleneck between the image encoder and the decoders. It uses a set of $N$ learnable query tokens to probe the encoded image features through cross-attention, producing $N$ refined query embeddings of dimension $d$ that capture the relevant visual information in a form amenable to language modeling. We refer to the set of vision-language query embeddings $\mathbf{y}$ as the \textbf{VLF}. In BLIP-2, $N=32$ queries of dimension $d=768$ are used, yielding a total feature size much smaller than the raw image or even the full set of patch embeddings (\emph{this compression ratio remains fixed even if the image resolution increases}). The transmitter directly maps $\mathbf{y}$ to analog complex symbols over the channel, avoiding autoregressive token generation or pixel/patch-domain source-channel coding~\citep{Liang_2025_CVPR}. This design keeps the transmitter lightweight while leveraging receiver-side priors for semantic alignment and reconstruction.

\subsection{Physical Channel}
The VLF $\mathbf{y}$ is transmitted over a wireless channel with additive noise. We model the channel as Rayleigh fading with additive white Gaussian noise (AWGN), a standard choice for dynamic wireless environments. The matrix $\mathbf{y}$ is treated as a sequence of complex symbols of length $N \cdot d /2$ (i.e., $\mathbf{y} \in \mathbb{C}^{N \cdot d/2}$). The received feature $\tilde{\mathbf{y}}$ is given by:
\begin{equation}
\tilde{\mathbf{y}} = \mathbf{h} \cdot \mathbf{y} + \mathbf{n},
\end{equation}
where $\mathbf{h} \in \mathbb{C}^{N \cdot d/2}$ is the element-wise channel gain and $\mathbf{n} \sim \mathcal{CN}(\mathbf{0}, \sigma^2 \mathbf{I}_{N \cdot d /2})$ the Gaussian noise. The effective SNR is $\gamma ={\bar{P} \cdot \mathbb{E}[|\mathbf{h}|^2]}/{\sigma^2}$, where $\bar{P}$ is the average transmit power per feature element. We note that by assuming $\mathbb{E}[|h_i|^2]=1$, $\gamma$ simplifies to $\bar{P}/\sigma^2$. Overall, this Rayleigh fading AWGN channel model provides a baseline for evaluating system performance under realistic wireless impairments, where each transmitted feature symbol is independently attenuated and corrupted by noise.

\paragraph{Bandwidth Compression Ratio} 

Bandwidth compression ratio (BCR) quantifies the spectral load of sending the VLF as channel uses per pixel. With the transmitted feature $\mathbf{y}\in\mathbb{C}^{N d/2}$ mapping to $n_{\text{ch}}=N d/2$ complex channel uses, we define
% Transmission efficiency is measured by the \emph{bandwidth compression ratio} (BCR)~\citep{deepjscc_q, diff_jscc}, defined as
\begin{equation}
\mathrm{BCR}=\frac{n_{\text{ch}}}{C \cdot h \cdot w}=\frac{N \cdot d}{2C \cdot h \cdot w}.
\end{equation}
% with input size $C\times h\times w$ ($C{=}3$ for RGB). Lower BCR indicates higher spectrum efficiency. Since $(N,d)$ are fixed, the VLF size remains constant regardless of input resolution, so BCR decreases as image size grows. This scalability highlights the efficiency of transmitting semantic features instead of pixel-level data.
for an input of size $C\times h\times w$ ($C{=}3$ for RGB). Since $(N,d)$ are fixed, $\mathrm{BCR}\propto 1/(h\,w)$ and thus decreases as resolution increases (e.g., with $N{=}32$ and $d{=}768$, the BCR for a $256{\times}256$ RGB image is $1/16$). Equivalently, the number of channel uses $n_{\text{ch}}$ is resolution-invariant, so our system keeps a uniform per-image transmit budget---bandwidth does not scale with pixel count, yielding higher spectral efficiency for high-resolution content.

\subsection{Multimodal Semantic Decoders}
At the receiver, the goal is to reconstruct text and/or image from the noisy VLF $\tilde{\mathbf{y}}$. The VLF-MSC receiver includes two semantic decoders: a text decoder that generates a caption and an image decoder that reconstructs the image. Both are conditioned on the same $\tilde{\mathbf{y}}$. The text decoder follows BLIP-2, feeding the VLF into a decoder-based LLM, while the image decoder is inspired by BLIP-Diffusion~\citep{blip_diffusion} to generate images aligned with $\tilde{\mathbf{y}}$. Unlike BLIP-Diffusion, our design avoids auxiliary prompts at inference, keeping the system practical in communication. Each decoder is detailed below.

\subsubsection{Text Semantic Decoder}
The text semantic decoder is a decoder-based LLM (e.g., OPT~\citep{zhang2022opt}) that generates $\hat{\mathbf{x}}_{\text{txt}}$ from received VLFs. The noisy VLF $\tilde{\mathbf{y}}$ is average-pooled and projected into the LLM embedding space:
$\mathbf{e}_0 = W_{\text{proj}}(\tfrac{1}{N}\sum_{i=1}^N \tilde{\mathbf{y}}_i)$.
This vector is prepended as a soft prompt, providing context without extra tokens. The LLM then autoregressively predicts tokens $\hat{o}_t$ until an end-of-sequence is reached, which are detokenized into $\hat{\mathbf{x}}_{\text{txt}}$.

\subsubsection{Image Semantic Decoder}
% The image semantic decoder generates $\hat{\mathbf{x}}_{\text{img}}$ from VLF alone. We implement it as a VLF-conditioned latent diffusion model (LDM), where the U-Net receives $\tilde{\mathbf{y}}$ via cross-attention at each denoising step to enforce semantic alignment.
The image semantic decoder generates an image $\hat{\mathbf{x}}_{\text{img}}$ from the received VLF as its sole guidance. We implement this branch as a VLF-conditioned latent diffusion model (LDM), inspired by BLIP-Diffusion~\citep{blip_diffusion} but without requiring text prompts at the receiver. The LDM operates in a perceptual latent space and iteratively denoises a latent variable toward the target image while being conditioned on $\tilde{\mathbf{y}}$. Specifically, the U-Net receives $\tilde{\mathbf{y}}$ through cross-attention at each denoising step, thereby maintaining semantic alignment between the generated image and the source image.

\paragraph{Training Strategy}
% During training, the LDM is conditioned on both the VLF and a subject label $c_i$ using the latent diffusion objective~\citep{ddpm}. The label is encoded by CLIP~\citep{clip}, while the VLF is projected into the same embedding space. The U-Net then predicts noise $\epsilon_\theta(\mathbf{z}_t,t\mid \mathbf{e}^{\text{text}}_i)$ given latent $\mathbf{z}_t$ at step $t$, optimized via mean squared error. This enables synthesis reflecting both high-level subject (label) and s VLF. At inference, only the VLF is used.
During training, the LDM is conditioned on both the VLF and a subject label $c_i$ to learn subject-driven generation, using the simplified latent diffusion objective~\citep{ddpm}. The label is tokenized and processed by the CLIP~\citep{clip} text encoder, while the image VLF is first projected by a learned linear map into the text-encoder embedding space and provided as context embeddings. The text encoder’s output sequence $\mathbf{e}^{\text{text}}_i$ subsequently conditions the U-Net via cross-attention at each denoising step. Given a latent variable $\mathbf{z}_t$ at diffusion timestep $t$, the conditional U-Net predicts the added noise $\epsilon_\theta(\mathbf{z}_t, t | \mathbf{e}^{\text{text}}_i)$, and the model is trained to minimize the mean squared error between the prediction and the true noise. Through this process, the diffusion model learns to synthesize images that reflect both the high-level concept (from the text label) and the semantic features (from the VLF) of the training samples.

\paragraph{Prompt-Free VLF-to-Image Generation}
At inference (communication phase), we generate images without any explicit text prompts, relying solely on the received VLF $\tilde{\mathbf{y}}$ as the condition. Notably, BLIP-Diffusion showed that providing a subject text to the multimodal encoder can moderately improve generation metrics. However, in practical semantic communication, transmitting an additional subject description would introduce significant bandwidth overhead and latency, making the system less viable for real-time applications. Therefore, our design deliberately avoids such auxiliary prompts, ensuring the framework remains lightweight and deployable. Concretely, we first project the received VLF $\tilde{\mathbf{y}}$ into the text encoder embedding space and pass it as context. The text encoder’s output sequence then conditions the U-Net via cross-attention at each denoising step, guiding the image generation. This enables the receiver to reconstruct a semantically consistent image $\hat{\mathbf{x}}_{\text{img}}$, using only the information contained in the projected VLF to faithfully represent the original image $\mathbf{x}$.

Unlike a conventional system that might transmit a compressed image or separately send both image and caption, our semantic approach conveys a unified representation through the VLF. The receiver’s decoders then regenerate the desired modality, effectively compressing the source information in a task-oriented manner. This communication-centric design highlights the key advantage of VLF-MSC: enabling multimodal reconstruction with minimal transmission burden while maintaining semantic integrity across modalities.

\section{Experiments}
\subsection{Experimental Details}
Our experiments were conducted on the Open Image V6 dataset~\cite{kuznetsova2020open}, with all images resized to $256 \times 256$ resolution. All models were evaluated over an AWGN channel with SNR levels ranging from $-5$~dB to $10$dB in 2.5~dB increments.
\paragraph{Baselines}
To assess the effectiveness of the proposed VLF-MSC framework, we compare it with three representative alternative approaches. Each baseline is a semantic communication method tailored to a single modality:
\begin{itemize}
    \item \textbf{VLF-MSC} (image, text): We adopt the BLIP-2 architecture with an EVA-CLIP~\citep{eva-clip} ViT-g/14 image encoder, a Q-Former with 32 learnable queries, and the OPT-6.7B language model as the text decoder. For text conditioning, we use a CLIP-based text encoder, and the semantic image decoder is implemented with Stable Diffusion v1.5~\cite{LDM}. The diffusion model was fine-tuned by randomly applying DDPM training steps sampled from $[0,100]$, and inference is performed with 50 DDIM steps. The BCR for a $256\times256$ input image is 1/8; for higher resolutions, the BCR scales inversely with the number of pixels.
    \item \textbf{Img2Img-SC}~\cite{LatentSC} (image): Img2Img-SC is a two-stage semantic communication framework. It transmits both an image latent representation and a textual caption (the caption is generated from the source image), while the latent helps preserve visual details. At the receiver, a pre-trained text-to-image generator (Stable Diffusion) reconstructs the image conditioned on the caption. The combined BCR is about 1/11 and remains roughly constant across image resolutions.
    \item \textbf{DeepSC}~\cite{DeepSC1} (text): DeepSC is a pioneering text-to-text neural semantic communication model that applies a Transformer-based encoder-decoder to textual data. It adopts a joint semantic-channel coding design, in which both source and channel coding are learned together within a unified neural framework. In our evaluation, each input image is first preprocessed by BLIP-2 into a textual description, which is then transmitted using the DeepSC. 
    \item \textbf{ASCII + (7,4) Hamming code} (text): We implement a traditional separate source-channel coding baseline. The input image is first converted to text using a BLIP-2 model (to produce a caption). The resulting text tokens are then encoded in 7-bit ASCII, partitioned into 4-bit blocks, and each block is encoded with a $(7,4)$ Hamming code. The bit stream is modulated using 16-QAM for transmission over the channel.
\end{itemize}

\paragraph{Evaluation Metrics}
% For text evaluation, we use BLEU~\cite{BLEU}, which measures n-gram overlap, and BERT score~\cite{BERT}, which assesses semantic similarity via contextual embeddings from a pre-trained BERT model. For images, we adopt LPIPS~\cite{Zhang_2018_CVPR} for perceptual similarity and CLIP score~\cite{clip} for image-text semantic alignment.
For evaluation, we employ two metrics for each modality. For the received text, we use the Bilingual Evaluation Understudy (BLEU) score~\cite{BLEU}, which calculates n-gram overlaps between generated and reference text, and BERT score~\cite{BERT}, which computes semantic similarity by comparing contextual embeddings from a pre-trained BERT model. Both metrics range from 0 to 1, and higher values indicate greater semantic similarity to the original text. For the received images, we use Learned Perceptual Image Patch Similarity (LPIPS)~\cite{Zhang_2018_CVPR} to measure the perceptual similarity between the transmitted and original images, and the CLIP score~\cite{clip} to quantify the semantic alignment between the received image and the original image’s caption. A lower LPIPS indicates higher perceptual fidelity, while a higher CLIP score signifies better image-text consistency.

\subsection{Experimental Results}

\begin{figure}[!t]
\begin{center}
\centerline{\includegraphics[width=0.995\linewidth]{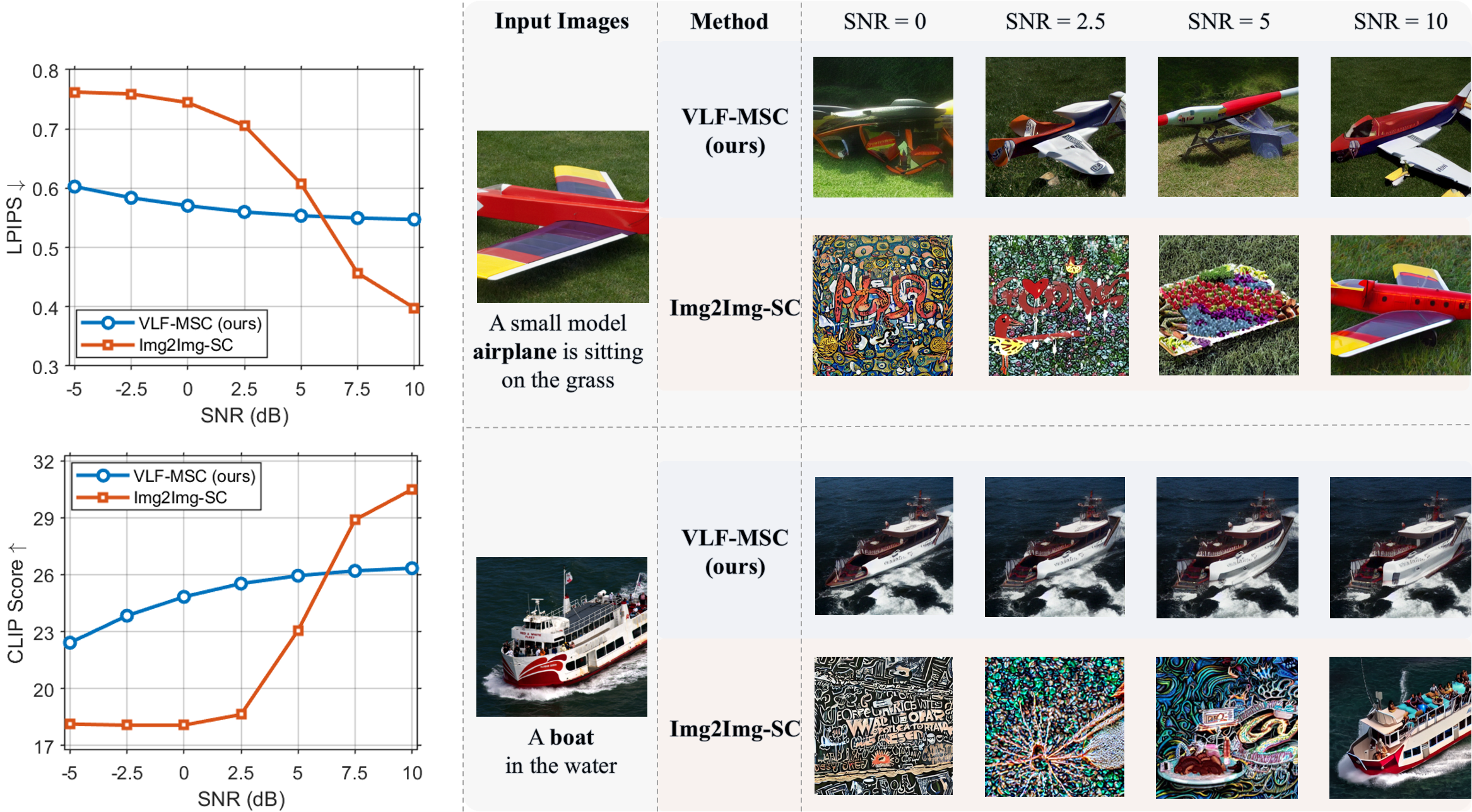}}
\caption{\textbf{Image transmission} results. 
\emph{Left}: LPIPS (perceptual similarity) and CLIP score (semantic alignment) across SNR levels. 
\emph{Right}: qualitative comparison of reconstructed images between VLF-MSC and Img2Img-SC.}
\vskip -0.2in
\label{fig:result_img}
\end{center}
\end{figure}

\paragraph{Image Transmission Performance}
We present quantitative and qualitative comparisons of VLF-MSC with the Img2Img-SC baseline in Figure~\ref{fig:result_img}. VLF-MSC achieves higher perceptual similarity (lower LPIPS) and semantic alignment (higher CLIP score) across SNRs from $-5$~dB to $5$~dB. This shows that the VLF representation remains robust both perceptually and contextually even under severe noise. Unlike conventional image transmission that directly feeds image latents to a decoder, our method converts the image into a VLF and transmits it via the text modality, which contributes to improved robustness. Similar trends appear in the qualitative results (Figure~\ref{fig:result_img}, right), where the proposed method produces outputs faithful to the original even at low SNRs. For example, the main objects (airplane, boat) remain robustly conveyed despite high noise levels. These results suggest that VLF-MSC goes beyond compression and reconstruction by transmitting a noise-resilient semantic representation that conventional methods cannot provide.

\begin{figure}[!t]
\begin{center}
\centerline{\includegraphics[width=0.995\linewidth]{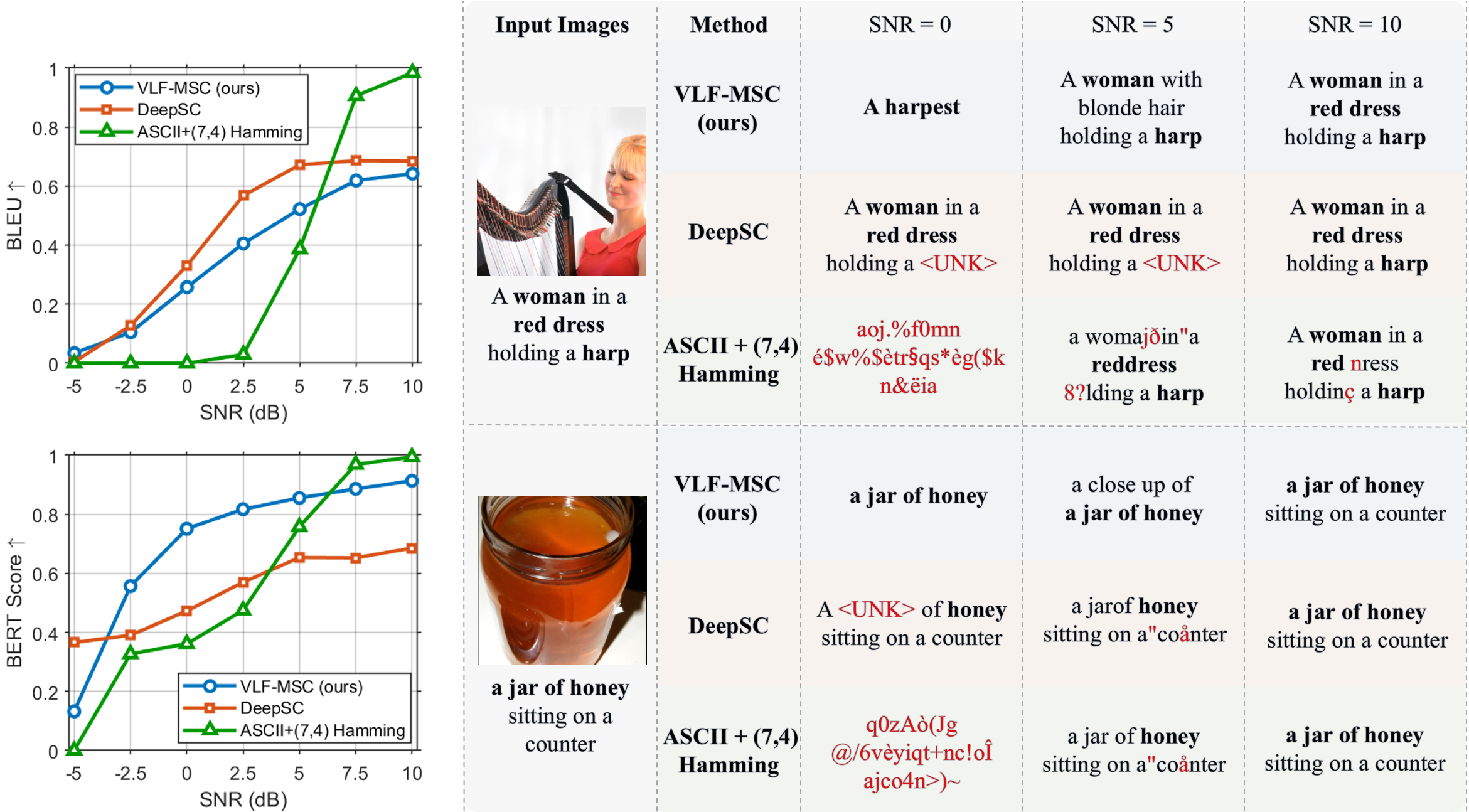}}
\caption{\textbf{Text transmission} results. 
\emph{Left}: BLEU (lexical overlap) and BERT score (semantic similarity) across SNR levels. 
\emph{Right}: qualitative comparison of reconstructed texts among VLF-MSC, DeepSC, and separation baseline.}
\vskip -0.2in
\label{fig:result_txt}
\end{center}
\end{figure}

\paragraph{Text Transmission Performance}
Figure~\ref{fig:result_txt} compares our VLF-MSC with the established text-based SC system DeepSC and the traditional ASCII+(7,4) Hamming code baseline. In terms of BLEU, which scores lexical overlap between sentences, DeepSC performs better since it follows a token-level reconstruction strategy. In contrast, VLF-MSC emphasizes semantic similarity, measured by BERT score, and consistently outperforms both baselines across all SNRs. In summary, by leveraging a semantically rich VLF, the proposed VLF-MSC framework enables robust text generation that maintains the linguistic structure and meaning of the source content. The experimental results show that even under varying channel conditions, our method is able to preserve semantic consistency and language integrity more effectively than conventional approaches. Additional qualitative results of joint image-text generation in VLF-MSC are provided in Appendix Figure~\ref{fig:app:joint}.

\section{Conclusion and Future Work}
In conclusion, we proposed VLF-MSC, a unified vision-language foundation model based semantic communication framework for joint image and text transmission. By leveraging state-of-the-art multimodal models (BLIP-2, Stable Diffusion) and LLMs, VLF-MSC achieves robust performance under noisy channel conditions, significantly improving semantic fidelity in both visual and textual domains at low SNRs. This work contributes to next-generation wireless communications by introducing a paradigm that transmits high-level semantic representations instead of low-level bit sequences, enhancing both reliability and efficiency. 

An important future direction is to perform more fine-grained ablation studies on the number of transmitted queries and feature dimensions, which directly impact both bandwidth consumption and semantic quality. Furthermore, exploring the bandwidth-quality trade-offs across varying image resolutions and message complexities will provide deeper insights into the scalability of the proposed framework. Beyond these analyses, further extending the performance gains to higher-SNR regimes and optimizing the framework for real-time and multi-user scenarios will further advance the practical deployment of semantic communications in next-generation networks.

\begin{ack}
This work was partly supported by the Institute of Information \& Communications Technology Planning \& Evaluation (IITP)-ITRC (Information Technology Research Center) grant funded by the Korea government (MSIT) (IITP-2026-RS-2020-II201787, contribution rate: 50\%) and (IITP-2026-RS-2023-00259991, contribution rate: 50\%)
\end{ack}

\bibliography{references}
\bibliographystyle{plainnat}

%%%%%%%%%%%%%%%%%%%%%%%%%%%%%%%%%%%%%%%%%%%%%%%%%%%%%%%%%%%%

\appendix
\newpage

\section{Additional Qualitative Results}

\begin{figure}[!h]
\centering
\includegraphics[width=0.99\linewidth]{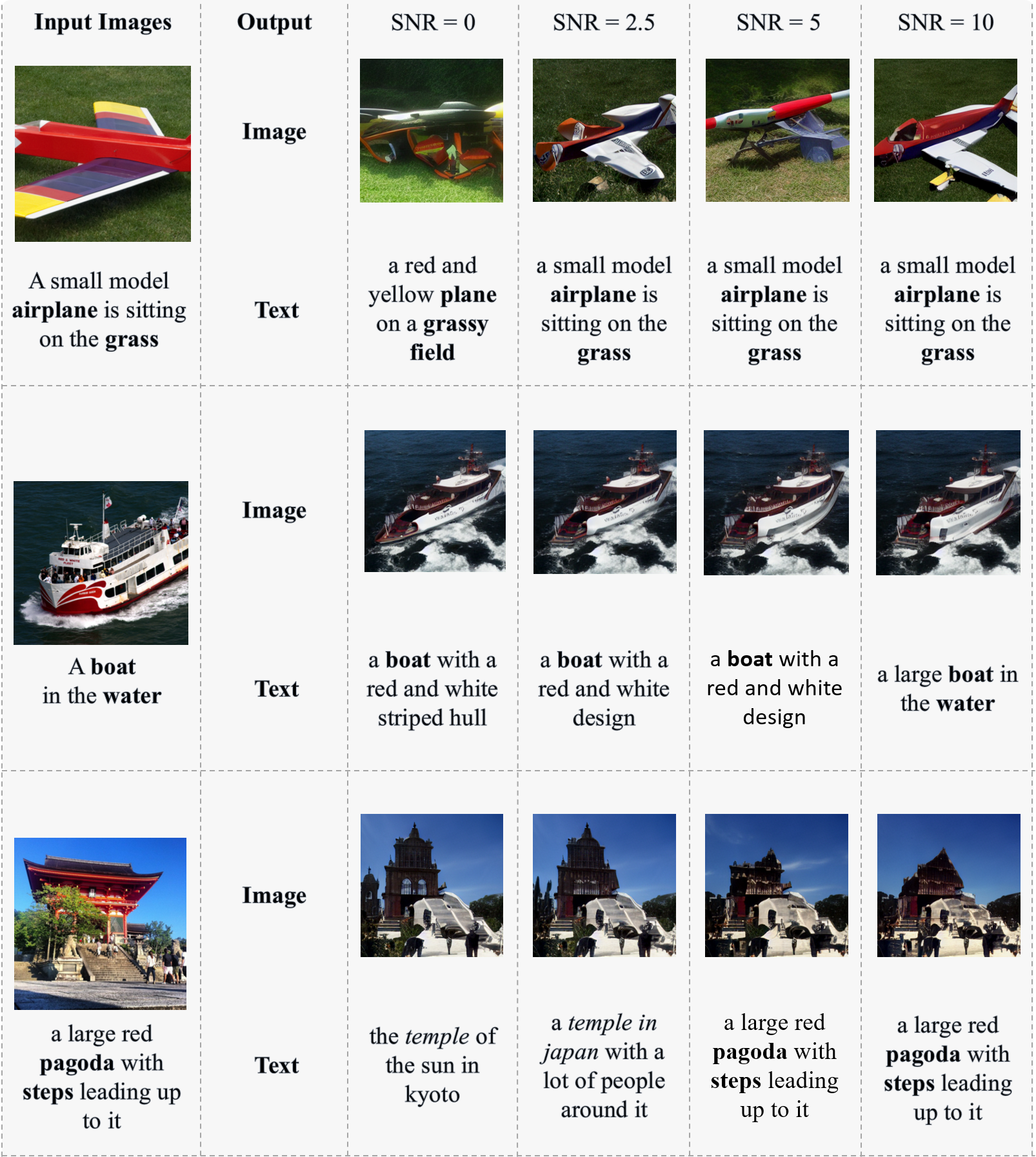}
\caption{Image-text semantic reconstruction under varying SNRs using the proposed VLF-MSC. 
A single transmitted VLF conditions both modalities, eliminating modality-specific streams and producing semantically aligned outputs across SNRs. 
\textbf{Bold} words denote preserved key semantics.}
\label{fig:app:joint}
\end{figure}

Fig.~\ref{fig:app:joint} shows additional joint results of image semantic reconstruction and text generation across channel conditions. For each input image (left), we transmit a single vision–language feature (VLF) once over the channel and condition both the semantic image decoder and the semantic text decoder on the same VLF at the receiver. Columns correspond to SNR $\in\{0,\,2.5,\,5,\,10\}$\,dB. As SNR increases, perceptual fidelity improves while object identity and scene semantics remain stable even at low SNR, illustrating cross-modal consistency enabled by a unified representation.

\end{document}